\documentclass[conference, 9pt]{IEEEtran}
\IEEEoverridecommandlockouts
\usepackage[colorlinks=true, linkcolor=black, urlcolor=black, citecolor=black]{hyperref}
\usepackage{cite}
\usepackage{amsmath,amssymb,amsfonts}
\usepackage{algorithmic}
\usepackage{graphicx}
\usepackage{textcomp}
\usepackage{xcolor}
\usepackage{makecell}
\usepackage{booktabs}
\usepackage{tipa}
\usepackage{tikz}

\def\BibTeX{{\rm B\kern-.05em{\sc i\kern-.025em b}\kern-.08em
    T\kern-.1667em\lower.7ex\hbox{E}\kern-.125emX}}

\begin{document}

\title{Sagalee: an Open Source Automatic Speech Recognition Dataset for Oromo Language\\
\thanks{\IEEEauthorrefmark{4} Corresponding Authors: Thomas Fang Zheng and Dong Wang. \\ This work was supported by the National Natural Science Foundation of China (NSFC) under Grants No.62301075.}
}

\author{
\IEEEauthorblockN{Turi Abu\IEEEauthorrefmark{1}\IEEEauthorrefmark{2}, Ying Shi\IEEEauthorrefmark{1}\IEEEauthorrefmark{3} , Thomas Fang Zheng\IEEEauthorrefmark{1}\IEEEauthorrefmark{4}  and Dong Wang\IEEEauthorrefmark{1}\IEEEauthorrefmark{4}}
\IEEEauthorblockA{\IEEEauthorrefmark{1} Center for Speech and Language Technologies, BNRist, Beijing
}
\IEEEauthorblockA{\IEEEauthorrefmark{2}Department of Computer Science and Technology, Tsinghua University, Beijing, China
}

\IEEEauthorblockA{\IEEEauthorrefmark{3} School of Computer Science and Technology, Harbin Institute of Technology, Harbin, China.}
\IEEEauthorblockA{
Email: \url{tur23@mails.tsinghua.edu.cn}
}

}

\maketitle

\begin{abstract}
We present a novel Automatic Speech Recognition (ASR) dataset  for the Oromo language, a widely spoken language in Ethiopia and neighboring regions. The dataset was collected through a crowd-sourcing initiative, encompassing a diverse range of speakers and phonetic variations. It consists of 100 hours of real-world audio recordings paired with transcriptions, covering read speech in both clean and noisy environments. This dataset addresses the critical need for ASR resources for the Oromo language which is underrepresented. To show its applicability for the ASR task, we conducted experiments using the Conformer model, achieving a Word Error Rate (WER) of 15.32\% with hybrid CTC and AED loss and WER of 18.74\% with pure CTC loss.  Additionally, fine-tuning the Whisper model resulted in a significantly improved WER of 10.82\%.  These  results establish baselines for Oromo ASR, highlighting both the challenges and the potential for improving ASR performance in Oromo. The dataset is publicly available at \href{https://github.com/turinaf/sagalee}{https://github.com/turinaf/sagalee} and we encourage its use for further research and development in Oromo speech processing.
\end{abstract}


\begin{tikzpicture}[remember picture, overlay]
  \node[anchor=south, yshift=10mm, text width=2\linewidth, align=center, 
        draw, font=\small] at (current page.south) {
    © 2025 IEEE. Personal use of this material is permitted. Permission from IEEE must be obtained for all other uses, in any current or future media, including reprinting/republishing this material for advertising or promotional purposes, creating new collective works, for resale or redistribution to servers or lists, or reuse of any copyrighted component of this work in other works.
  };
  \end{tikzpicture}
\section{Introduction}
\label{sec:intro}
Automatic Speech Recognition (ASR) technology has witnessed remarkable progress, revolutionizing human-computer interaction across diverse domains \cite{Kheddar_2024, malik2021automatic}. However, this progress has not been uniformly distributed across all languages. Many low-resource languages, particularly those spoken in developing countries, remain significantly underrepresented in ASR research and development. This issue becomes more critical in the deep learning era, where the core component of ASR systems is formed by large and complex neural networks, which require a large amount of data for the model training. 

The scarcity of resources for these languages encompasses several critical aspects. First, there is data scarcity, where the availability of large, transcribed speech datasets, crucial for training accurate ASR models, is severely limited. Without sufficient data, developing reliable speech recognition systems becomes a significant challenge. Second, linguistic resources such as pronunciation dictionaries, grammar, and other tools that aid in understanding the structure and nuances of a language are often underdeveloped or nonexistent. This lack of foundational resources makes it difficult to build systems that can accurately process and interpret these languages. Third, there are acoustic modeling challenges, as the acoustic characteristics of under-resourced languages may differ significantly from those well-represented in existing models, requiring specialized techniques to address these differences. This disparity not only limits the accessibility of speech-based technologies for millions of speakers but also hinders the preservation of linguistic diversity in the digital age, risking the loss of unique cultural and linguistic heritage.

The Oromo language, spoken by over 45 million people primarily in Ethiopia and neighboring countries \footnote{https://en.wikipedia.org/wiki/Oromo\_language}, exemplifies this challenge. Despite being one of the most widely spoken languages in Africa, Oromo suffers from a severe scarcity of resources crucial for building robust ASR systems. 
This scarcity encompasses various aspects, including the availability of transcribed speech data and linguistic resources like pronunciation dictionaries. To the best knowledge of the authors, there are two public speech datasets, one involves 3hrs of speech from 6 speakers\footnote{https://data.mendeley.com/datasets/hnvkvj589y/1} and the other containing 17hrs of read speech from a single male speaker\footnote{https://data.mendeley.com/datasets/mpy85ns82z/2}. Both the datasets were designed for text-to-speech synthesis and none of them can be used for developing ASR system, due to the limited number of speakers and the constrained recording environment. 

The lack of resources hinders the development of essential applications such as voice assistants, dictation software, and language learning tools for Oromo speakers, further exacerbating the digital divide.
Addressing this gap, we introduce Sagalee, a novel and publicly available ASR dataset designed for the Oromo language. ``Sagalee" translates to ``voice" in Oromo, signifying the dataset's aim to give a voice to the language in the digital realm. This dataset, collected through a crowd-sourcing initiative facilitated by a dedicated mobile application, encompasses 100 hours of read speech data from 283 speakers. The use of a mobile application allowed us to reach a diverse pool of speakers across different geographical locations and demographic backgrounds, resulting in a dataset rich in phonetic and acoustic variations. This variety ensures the dataset's suitability for training and evaluating ASR systems capable of handling real-world scenarios.

This paper details the development of the Sagalee dataset, outlining its composition, the methodology employed for data collection and the measures taken to ensure data quality and diversity. We further present initial ASR experiments conducted using the Sagalee dataset. These experiments include training Conformer \cite{gulati2020conformer} from scratch and fine-tuning whisper \cite{radford2022whisper} large-v3 pretrained model. Our preliminary results achieve WER of 15.32\% for the Conformer model using a hybrid loss function combining Connectionist Temporal Classification (CTC) and Attention-based Encoder-Decoder (AED), and 18.74\% for the Conformer model trained with pure CTC loss. Finetuning whisper largev3 model achieved WER of 10.82\%. These results highlight the inherent challenges posed by low-resource language ASR while simultaneously demonstrating the viability of Sagalee dataset in enabling Oromo speech recognition research.


By making Sagalee publicly available, we aim to empower researchers and developers to advance Oromo speech technologies.

\section{The Oromo Language}

The Oromo language, known as `Afaan Oromoo' in native tongue, is a member of the Cushitic branch of the Afroasiatic language family\cite{tegegne2016development}. It is predominantly spoken by the Oromo people, who constitute the largest ethnic group in Ethiopia. It is also spoken by smaller communities in Kenya, Somalia, and diaspora population in other regions. With over 45 million speakers, it is one of the most widely spoken languages in Africa \footnote{https://en.wikipedia.org/wiki/Oromo\_language}. In Ethiopia, Oromo language is the official working language of the Oromia regional state government. 

\subsection{History}

The Oromo language has long history of oral traditions \cite{Demie1996HistoricalCI}. Despite its widespread use in the Horn of Africa and the significant number of speakers, it remained unwritten for a considerable period of time. 
This may be attributed to the Oromo people's traditional way of life \cite{OromoOrtho}. As a predominantly pastoral society, they relied heavily on oral communication for daily activities and passed down knowledge to future generations through spoken words.
There have been many attempts to transition Oromo from oral to written. The early attempt of writing Oromo goes back to the beginning of 19th century \cite{tegegne2016development}. The first script used to write was Arabic \cite{Demie1996HistoricalCI} which came to existence through religious introduction to Oromo people. However, the incompatibility of the Arabic script with the Oromo language is thought to have hindered the development and spread of written Oromo in the 19th century \cite{tegegne2016development}. In 1956, Shaykh Bakri Sapalo developed the Sapalo script, an indigenous writing system for the Oromo language, which represented its phonological features \cite{OromoOrtho}. However, this script wasn't widely adopted due to the lack of institutional support and infrastructure to promote its use\cite{OromoOrtho}. Moreover, unfavorable political conditions for writing, teaching, or broadcasting in Oromo until the 1970s further slowed the development of written Oromo \cite{Hassen1994SomeAO}. 

\subsection{Linguistic Characteristics}
\subsubsection{Writing System}
Today, the Oromo language primarily uses the Latin alphabet, known as Qubee. Qubee consists of the 26 letters of the standard Latin alphabet (A-Z) and digraphs knows as ``Qubee Dachaa" in Oromo to represent unique Oromo sound. Qubee was officially adopted as the writing system for Oromo in 1991, following years of advocacy for a standardized orthography that could accommodate the phonetic nuances of the language \cite{tegegne2016development, Gamta1993QubeAO}. Prior to this, Oromo was written using various scripts, including Arabic (early start of written Oromo) and Ge'ez script, which is still used for Amharic and Tigrinya, but these systems were less suited to the phonology of the Oromo language \cite{tegegne2016development, Gamta1993QubeAO}.

\subsubsection{Phonology}
Oromo has a rich phonological system that includes both consonants and vowels. It features five vowel sounds (a, e, i, o, u), which can be short or long, leading to a total of ten distinct vowel phonemes. Table \ref{tab:vowels} presents the vowels in Oromo organized by their articulation based on tongue position and the degree of mouth openness. The distinction between short and long vowels can alter the meaning of words. For example, ``lafa" means ``land, ground, earth," while ``laafaa" means ``soft." Table \ref{tab:phones} shows consonants in the Oromo language, categorized by place of articulation  and manner of articulation. The consonants p, v, and z in Oromo occur only in foreign words, i.e., words borrowed from other languages \cite{stroomer1987comparative, griefenow2001grammatical}.
\begin{table}[htb]
    \centering
    \caption{Vowels in Oromo language}
    \label{tab:vowels}
    \begin{tabular}{lcccc}
    \toprule
   & \textbf{Front} & \textbf{Central} & \textbf{Back} \\
    \midrule
        Closed &i /\textipa{I}/, ii /i:/ & & u /\textipa{U}/, uu/u:/  \\
        Mid & e /\textipa{E}/, ee/e:/ & & o /\textipa{O}/, oo/o:/ \\
        Open & & a /\textturna/, aa /\textipa{A}:/ & \\
    \bottomrule
    \end{tabular}
\end{table}

\begin{table}[htb!]
    \centering
    \caption{Consonants in Oromo language \cite{griefenow2001grammatical}}
    \label{tab:phones}
    \resizebox{\linewidth}{!}{
    \begin{tabular}{lccccccc}
    \toprule
        & & Labial & \makecell{Alveolar/ \\ Dental}  & Palatal & Velar & Glottal  \\
    \midrule
        Stop & Voiced & b & d & j & g  & \\
        Stop & Voiceless & (p) & t & ch & k & ' \\
        Stop & Glottalized & ph & x & c & q & \\
        Implosive & Voiced &  & dh &  &  &  \\
        Fricative & Voiced & (v) & (z) &  & &  \\
        Fricative & Voiceless & f & s & sh & & h \\
        Nasal & & m & n & ny & & \\
        Lateral & &  l, r & & & \\
        Glide & & w & & y & & \\
        
        \bottomrule
    \end{tabular}
     }
\end{table}

\subsubsection{Morphology}
Oromo is an agglutinative language, meaning it forms words and expresses grammatical relationships through the addition of prefixes and suffixes. This allows for a high degree of inflection and derivation. Nouns in Oromo can be marked for number, gender, and case, while verbs are conjugated to indicate tense, aspect, mood, and agreement with the subject. Example: Farda (singular) - horse, Fardeen (plural) - horses; mana (singular) -house , manoota (plural) houses.  The suffixes -een and -oota indicate number. Common suffixes in Oromo include: -oota, -ooli, -wwan, -lee, -an, -een, -eeyyii, -oo, -dha, -dhaan, -tiin, -tti, -ttii, -dhaaf.

\section{Literature Review}
\label{sec:literature}

\subsection{ASR for Low-Resource Languages}

Several approaches have been explored to overcome the low-resource challenges. One approach involves leveraging data and resources from high-resource languages to bootstrap ASR systems for low-resource languages. This can be achieved through techniques such as cross-lingual transfer learning \cite{yadav-sitaram-2022-survey}, where models pre-trained in high-resource languages are adapted to low-resource languages using limited data. Other commonly used methods include model distillation \cite{seth2023, yang2023knowledge}, self-supervised pre-training~\cite{baevski2019effectiveness}, data augmentation \cite{Park_2019, nguyen2020improvingseq2seqasr, Lam_2021}, and multi-task learning~\cite{chen2014joint}.
\begin{figure}[htb]
    \centering
    \includegraphics[width=1\linewidth]{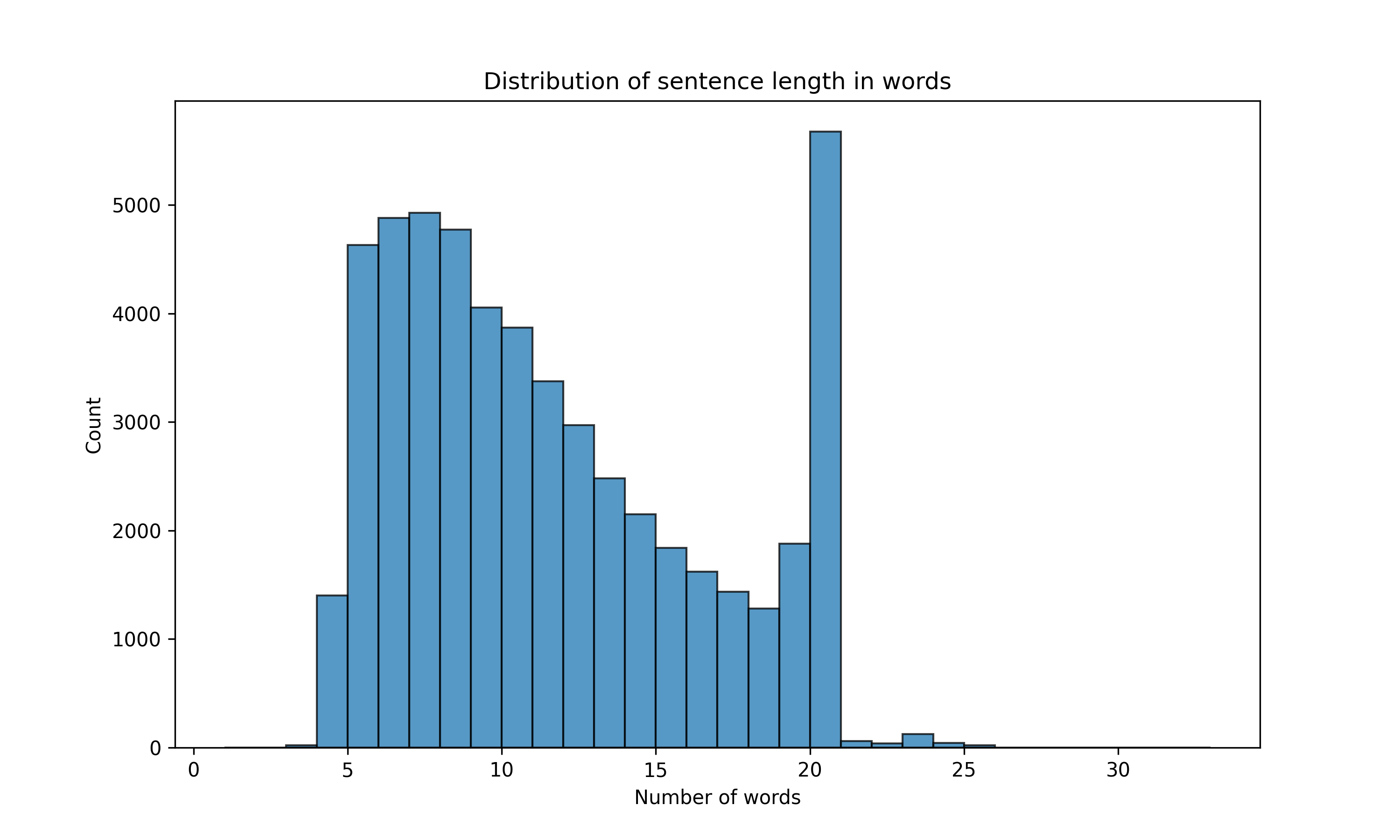}
    \caption{Distribution of Sentences Length in Number of Words}
    \label{fig:wordd}
\end{figure}
\subsection{Existing Work on Oromo ASR}

Despite its large speaker population, the Oromo language remains severely under-resourced in terms of ASR research. While some efforts have been made to develop resources, such as text corpora for machine translation that include 42K English-Oromo pairs \cite{chala2021}, very few studies have specifically focused on Oromo speech processing. Some preliminary work includes acoustic modeling for Oromo using Hidden Markov Models (HMMs) with limited data \cite{teshite2023afan} for 64 command words. Another ASR research was reported in~\cite{abate-etal-2020-large} using 22 hours of Oromo speech. All the datasets are small and not publicly available. 

\begin{figure}[htb]
\begin{minipage}[b]{.5\linewidth}
  \centering
  \centerline{\includegraphics[width=4.0cm]{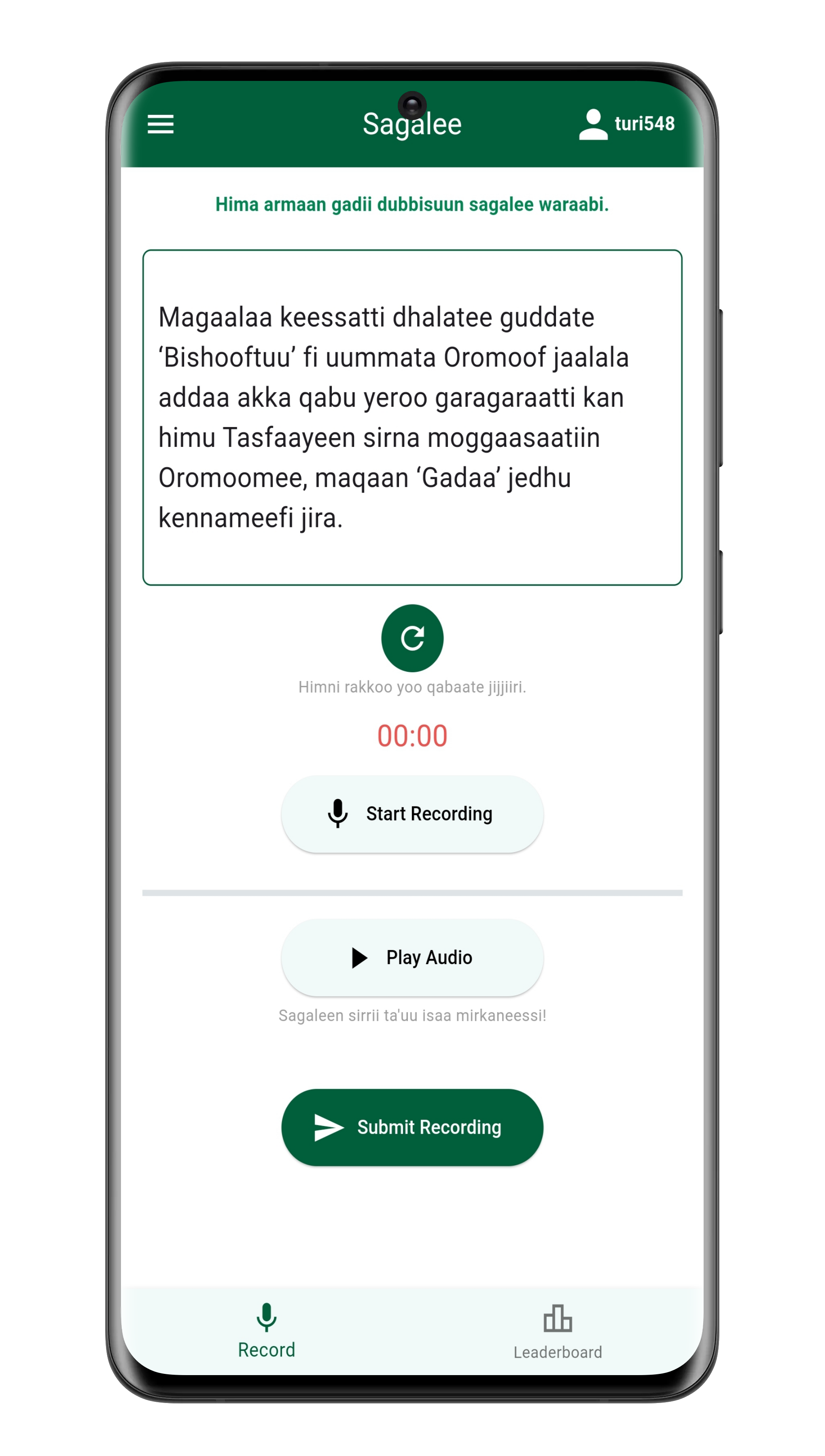}}
  \centerline{(a) Record}\medskip
\end{minipage}
\hfill
\begin{minipage}[b]{0.48\linewidth}
  \centering
  \centerline{\includegraphics[width=4.0cm]{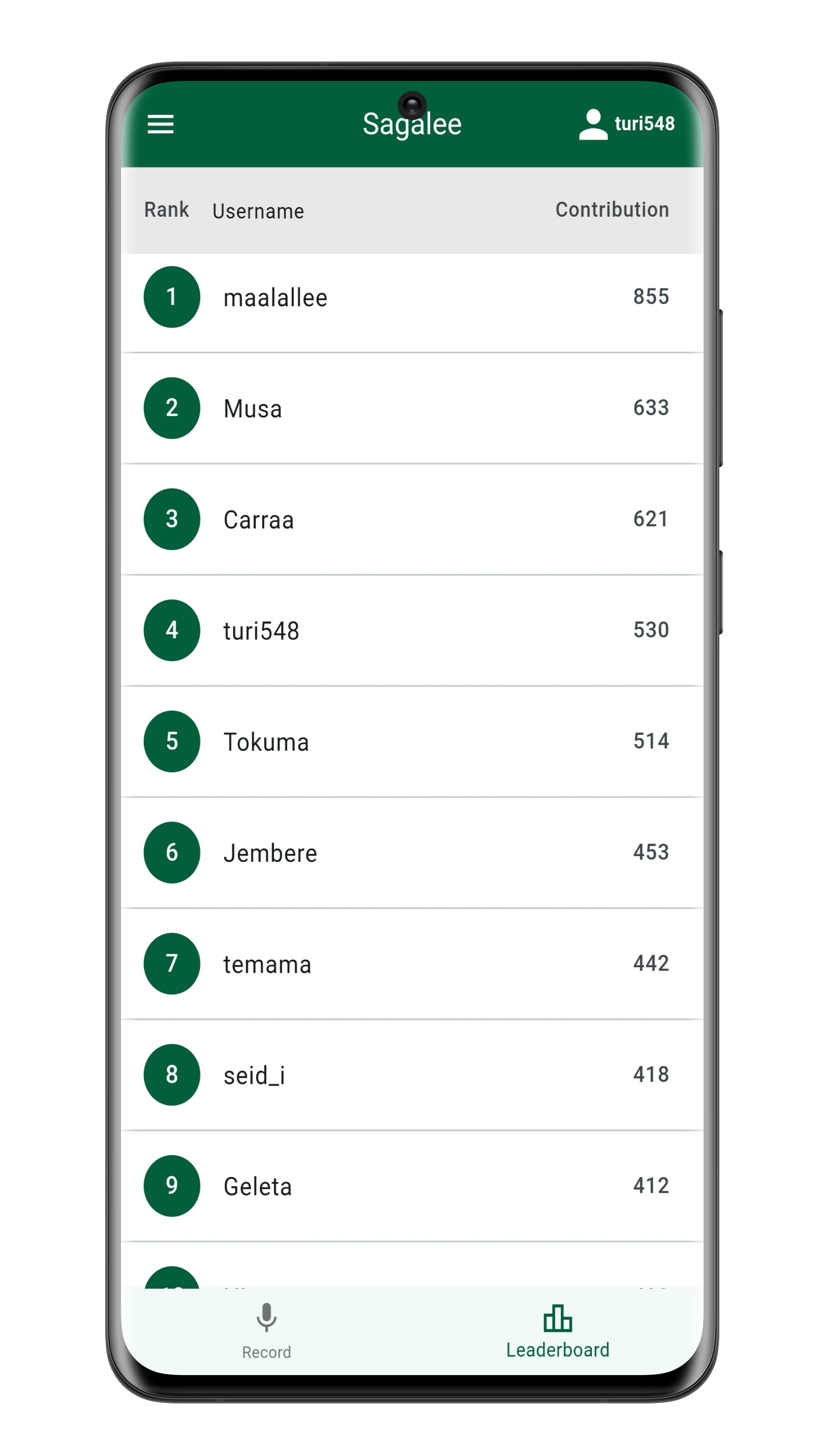}}
  \centerline{(b) Leaderboard }\medskip
\end{minipage}
\caption{Mobile App Developed for Data Collection.}
\label{fig:app}
\end{figure}

\section{Dataset Construction}
\label{sec:data}
\subsection{Speech Data Collection}
\subsubsection{Transcription Preparation}
Our data collection approach involved preparing transcriptions from various online sources including news articles, publicly available books on different topics such as fiction, history, politics and the Oromia regional state constitution. From these resources, we curated sentences with a minimum length of four words and a maximum of twenty five words, resulting in approximately 30,000 unique sentences. 

The length of sentences were limited to reduce reading difficulties for speakers. Figure \ref{fig:wordd} shows the distribution of the sentences in terms of word length. As we can see, majority of the sentences are 20 words long, mainly because we cut long sentences in documents to be 20 words long or less. Later we replaced some problematic sentences with sentences 20-25 words long.  
\subsubsection{Audio Recording}
We developed an Android app for mobile device that allows users to register an account and contribute their speech by reading the prepared sentences. 
On the developed Android app users can view sentences and record audio of the displayed text as shown in figure \ref{fig:app} (a). During registration, we collect demographic information, such as gender, age group, and dialect, which helps us better understand speakers. The audios are recorded using Android devices of the users with mono channel, sample rate of 24,000  and 16 bit rate. There is no restriction on the recording environment, as we hope to collect speech from diverse acoustic conditions. 
The sentences are presented to users in a randomized order. To ensure quality, we provided the speakers with a button to skip any sentence with spelling errors or hard to read. Contributors are compensated for their contributions. 

\subsection{Data Statistics}

\begin{figure}[htb]
\begin{minipage}[b]{.48\linewidth}
  \centering
  \centerline{\includegraphics[width=4.0cm]{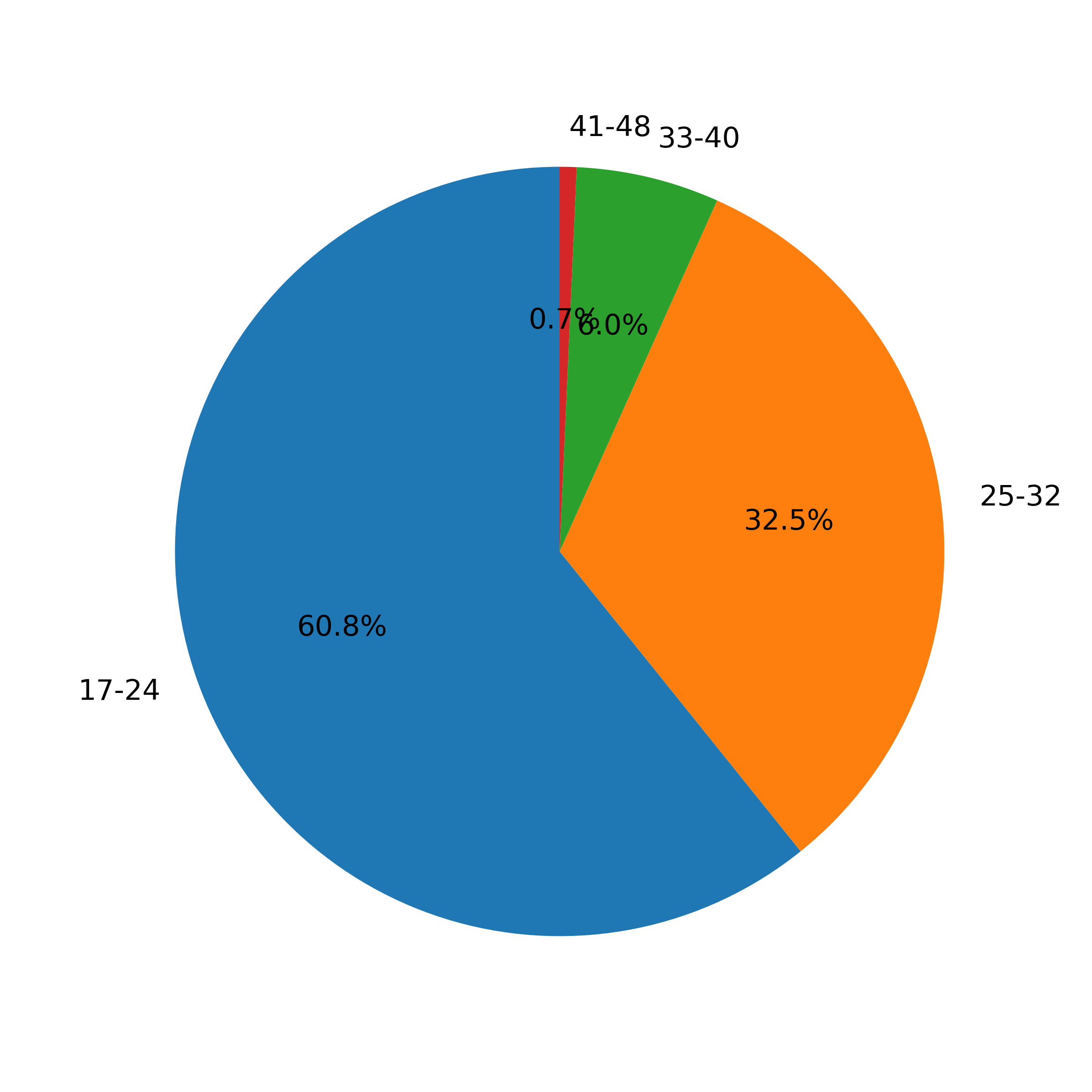}}
  \centerline{(a) Age groups of speakers}\medskip
\end{minipage}
\hfill
\begin{minipage}[b]{0.48\linewidth}
  \centering
  \centerline{\includegraphics[width=4.0cm]{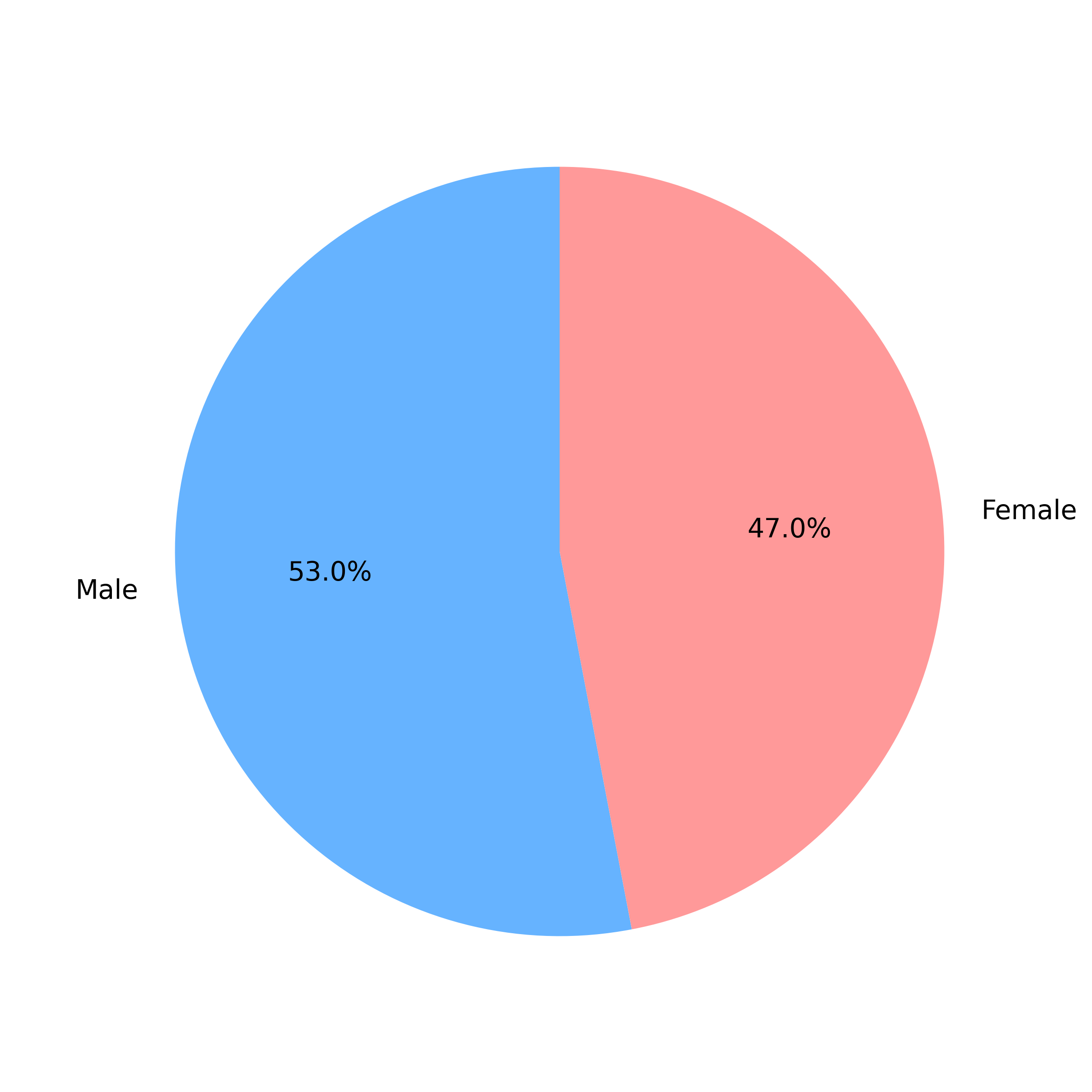}}
  \centerline{(b) Gender Distribution}\medskip
\end{minipage}

\begin{minipage}[b]{.48\linewidth}
  \centering
  \centerline{\includegraphics[width=4.0cm]{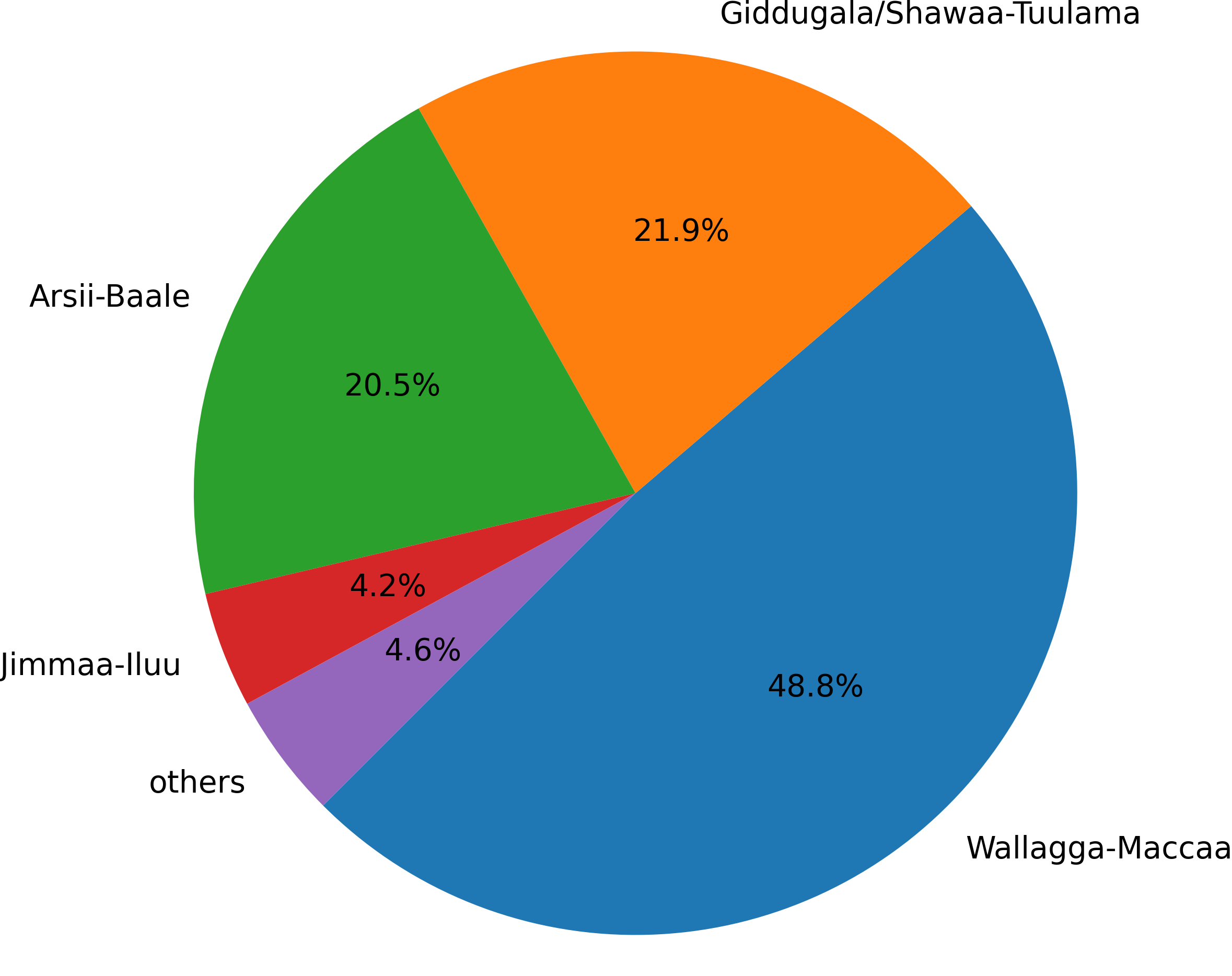}}
  \centerline{(c) Dialect distribution}\medskip
\end{minipage}
\hfill
\begin{minipage}[b]{0.48\linewidth}
  \centering
  \centerline{\includegraphics[width=4.0cm]{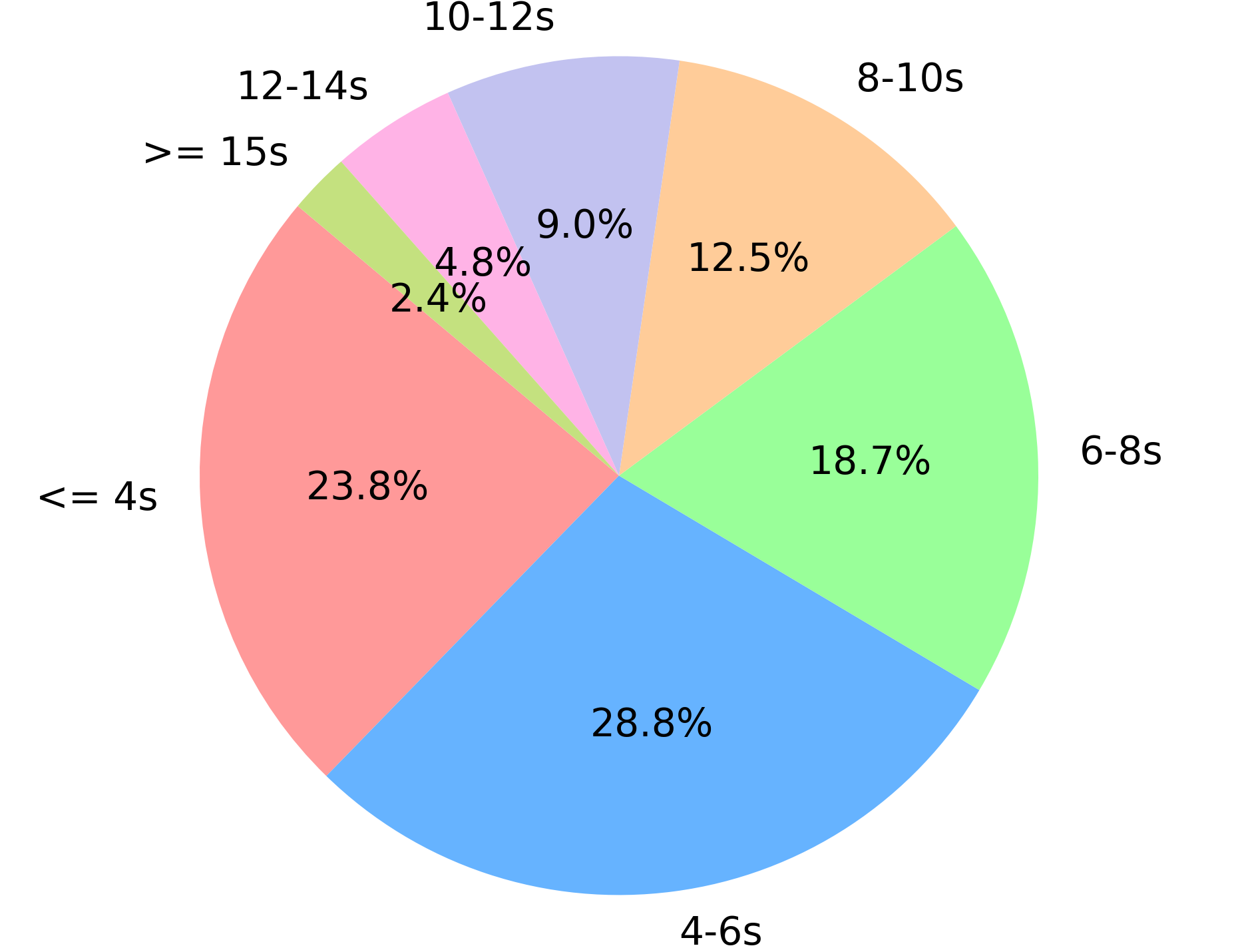}}
  \centerline{(d) Audio Duration in seconds }\medskip
\end{minipage}

\caption{Distribution of Sagalee dataset across age groups, dialect, gender and duration}
\label{fig:stats}
\end{figure}

Table~\ref{tab:summary} presents the basic statistics of the Sagalee dataset. 
The pie chart in figure \ref{fig:stats} (a) indicates the percentage of speakers' age group.  
There are 283 speakers in total out of which 150 are male and 133 are female speakers as indicated in Figure \ref{fig:stats} (b). 

Regarding the dialect, we asked the speakers to select their dialects based on their geographical origin (birth place), since there are no widely accepted or consistent categories of Oromo dialects 
 familiar to the general population \cite{tegegne2016development}, where zones with same dialect are together for example the central shaw zones, the two Arsi and Bale. The options for the user to select were (1) Wallaggaa-Maccaa, (2) Giddugala/shawaa-Tuulama, (3) Arsii-Baalee, (3) Jimmaa-Iluu, (4) Harargee, (5) Boorana-Gabra, (6) Gujii, (7) Walloo-Raayyaa. 
 
 Finally, Figure \ref{fig:stats} (c) shows distribution of speakers from these categories. The audio duration is shown in a pie chart in Figure \ref{fig:stats} (d). Majority of the audios have duration between four seconds and six seconds. 
 \begin{table}[htb]
    \centering
    \caption{Summary of the Sagalee Dataset}
    \label{tab:summary}
    \resizebox{\linewidth}{!}{
    \begin{tabular}{|c|c|c|c|c|c|}
    \hline
    \makecell{Number of\\ speakers}  & \makecell{Number of\\Utterances} & \makecell{Avg Utterance \\ per speaker}  & \makecell{Avg\\ duration} & Avg Age & \makecell{Total\\duration}\\
    \hline
     283 & 53,573 & 192 &  6.7 sec &  24 & 100 hrs\\
     \hline
    \end{tabular}
    }
\end{table}

\section{EXPERIMENT}
\label{sec:typestyle}
We conducted experiments on the dataset we developed to show its suitability for the ASR task and establish the ASR baseline for the Oromo language. We built three models: (1) Conformer model trained from scratch using Sagalee, with the CTC-AED hybrid loss; (2) Conformer CTC model trained from scratch using Sagalee, with the solely CTC loss; (3) Whisper Largev3 model adapted from a pre-trained multi-lingual Whisper using Sagalee dataset. 

WeNet toolkit\footnote{https://github.com/wenet-e2e/wenet} \cite{yao2021wenet}, an end-to-end speech recognition framework, was used to perform the training and adaptation. The dataset is split into train, dev and test sets with 93.6hrs, 4.2hrs, and 2.4hrs size respectively. 

\subsection{Training From Scratch}
To build the Oromo ASR model from scratch, we employed the Conformer architecture \cite{gulati2020conformer}, which integrates convolutional and transformer \cite{vaswani2023attentionneed} layers to enhance performance in speech processing tasks. The Wenet toolkit was used for training, leveraging its two-pass approach with a hybrid CTC and AED loss function \cite{yao2021wenet} given in equation \ref{eq:loss}
\begin{equation}
\text{Loss}_{\text{hybrid}} = \lambda  \text{L}_{\text{CTC}}(x,y) + (1 - \lambda) \text{L}_{\text{AED}}(x, y)
\label{eq:loss}
\end{equation}
where $\lambda$ is the CTC loss weight, $x$ is the acoustic feature and $y$ is the corresponding label.
For conformer CTC, we simply set the weight of CTC to 1 so that the loss is pure CTC loss. 
The encoder uses the Conformer architecture,  featuring an output size of 256 and four attention heads. It comprises 12 layers and applies a dropout rate of 0.1 to prevent overfitting. The input layer is a conv2d and includes a Convolutional Neural Network (CNN) module with a kernel size of 15. The activation function used is Swish \cite{ramachandran2017swish}, and it employs relative positional encoding. Layer normalization is applied before each block. The decoder is based on the Transformer architecture \cite{vaswani2023attentionneed}, configured with 6 layers and four attention heads, similar to the encoder. It also uses a dropout rate of 0.1 for regularization.

The tokenizer uses Byte Pair Encoding (BPE) \cite{sennrich2016bpe} in unigram mode. We first experimented with a large nbpe (number of subwords units) value of 5000, which led to excessive \texttt{<unk>} tokens and poor model performance. After experimenting with smaller values (50, 100, 500), we found nbpe of 100 and 500 significantly outperformed 50, with 500 yielding the best results. The results reported in this paper are based on nbpe of 500. 
We trained the model for 200 epochs with the above configuration. The testing for Conformer AED and Conformer CTC are performed using the average of 10 best checkpoints. 

\subsection{Adaptation From Whisper}
We fine-tuned the Whisper \cite{radford2022whisper} large-v3 model with 1.55 billion parameters using our dataset. The model employs a Transformer architecture for both the encoder and decoder, which share the same structural configuration. Specifically, both the encoder and decoder are composed of 32 Transformer blocks, each with 5120 hidden units and 20 attention heads. Additionally, we randomly initialized the convolutional layers at the front end of the encoder. For the encoder’s CTC projection layer, we initialized it with the weights from the word embeddings used in the decoder.
During fine-tuning, all parameters of the Whisper model were adjusted using gradients. The fine-tuning process employed a batch size of 2 and a learning rate of 1e-5. A warm-up learning rate scheduler was also used, with warm-up steps set to 12,000. The fine-tuning procedure lasted for a total of 15 epochs, and the final evaluation was based on the average of the model checkpoints from the last three epochs.

\subsection{Result}
\label{sec:result}
The performance of the trained models in terms of Word Error Rate (WER) is summarized in Table \ref{tab:result}. First, we observe that the Conformer AED and Conformer CTC models achieve WERs of 15.32\% and 18.74\%, respectively. These results are consistent with the experience in the ASR domain, that training-from-scratch model with low resource language of around 100 hours of training data may reach around 20\% WER \cite{damania22_interspeech} if the acoustic environment involves real complexity.

The Whisper model achieves a significantly lower WER, which is expected given its pre-training on a vast and diverse multilingual speech dataset. This pre-training enables the model to generalize effectively across languages, including low-resource ones like Oromo. Overall, the experimental results highlight the inherent challenges of the Oromo language, such as its unique phonetic and acoustic features, which can be difficult for models trained from scratch on limited data to address effectively. Importantly, these findings also underscore the efficacy of modern large-scale pre-training in tackling such challenges. Additionally, the results demonstrate the  value of the presented Sagalee dataset for advancing Oromo ASR.
 
\begin{table}[htb]
    \centering
    \caption{Experiment result on Sagalee dataset}
    \label{tab:result}
    \resizebox{\linewidth}{!}{
    \begin{tabular}{lcccc}
    \toprule
    \textbf{Model} & \textbf{Mode}  &  \textbf{WER \% } \\
    \midrule
        Conformer AED & From scratch  & 15.32 \\
        Conformer CTC & From scratch  & 18.74 \\
         Whisper Largev3 & Fine-tuning  &  10.82 &\\
    \bottomrule
    \end{tabular}
    }
\end{table}

\section{CONCLUSION AND FUTURE WORKS}
\label{sec:conclusion}
In this paper, we introduced Sagalee, an ASR dataset for the Oromo language, which aims to address the lack of resources for speech processing in the Oromo language. To show that the dataset is suitable for the ASR task, we have performed an experiment on the dataset using conformer architecture and also fine-tuning on pre-trained model, establishing a baseline ASR model. 
The baseline result shows the dataset can be used to train a reasonable ASR model with modern techniques. 
The future works will be expanding the dataset in terms of diversity and volume, and investigating its application in other speech processing tasks, such as speech synthesis and speaker recognition. 

\vfill\pagebreak

\end{document}